\documentclass[runningheads]{llncs}

\usepackage{eccv}

\usepackage{eccvabbrv}
\usepackage{graphicx}
\usepackage{booktabs}
\usepackage{multirow}
\usepackage{array}
\usepackage{amsmath}
\usepackage{amssymb}
\usepackage{xcolor}
\usepackage[accsupp]{axessibility}
\usepackage[breaklinks,colorlinks,citecolor=eccvblue]{hyperref}
\usepackage{subcaption}

\newcommand{\dataset}{\textsc{BD-HazardVLM}}
\newcommand{\qwen}{Qwen2.5-VL-7B}
\newcommand{\internvl}{InternVL3-8B}
\newcommand{\minicpm}{MiniCPM-V-4.6}

\begin{document}

\title{BD-HazardVLM: Probing Vision-Language Models for
Latent Defensive-Driving Hazards in Bangladesh Road Scenes}
\titlerunning{BD-HazardVLM: Defensive-Driving Hazards in Bangladesh Road Scenes}

\author{Abdullah Al Maruf\inst{1}\thanks{Corresponding author.} \and
Md. Sajedul Islam\inst{1} \and
Tanmoy Mridha\inst{1} \and
Irfan Hossain Bhuiyan\inst{1}}
\authorrunning{A. Al Maruf et al.}
\institute{Department of Computer Science and Engineering,\\
Rajshahi University of Engineering \& Technology (RUET), Rajshahi, Bangladesh\\
\email{\{2103115,2103096,2103118,2103088\}@student.ruet.ac.bd}}

\maketitle

\begin{abstract}
Vision-language models are increasingly used for driving-scene
understanding, yet their defensive reasoning under weakly structured
Global South traffic remains unclear. We introduce \dataset{}, a
diagnostic benchmark of 500 Bangladesh road-scene images---375 from
RSUD20K and 125 from TFP-BD---annotated for hazard presence, primary
hazard category, risk timing, severity, defensive actions, visible
evidence, and hallucination- and Western-road-bias traps. We evaluate
\qwen{}, \internvl{}, and \minicpm{} using a shared zero-shot task definition and structured output schema. Binary hazard-presence accuracy ranges from 88.0\% to 89.2\%,
essentially matching the 89.0\% majority-class baseline. Fine-grained
reasoning is substantially weaker: category accuracy remains below 40\%,
risk-timing accuracy ranges from 1.6\% to 12.4\%, and exact
defensive-action matching remains below 2\%. Although \internvl{} achieves
the highest severity accuracy at 61.2\%, it has zero recall on the 58
high-severity cases. Qwen exhibits severe category mode collapse, while all models show weak
latent-hazard anticipation and near-zero self-reported trap recognition. For the
evaluated models, these findings show that coarse hazard recognition does
not imply reliable fine-grained defensive-driving reasoning and motivate
geographically diverse safety evaluation.

\keywords{
Vision-language models
\and Defensive driving
\and Hazard reasoning
\and Bangladesh road scenes
\and Geographic robustness
}
\end{abstract}

\section{Introduction}
\label{sec:introduction}

Vision-language models (VLMs) are increasingly used for interpretable
driving-scene understanding, visual question answering, planning, and
language-guided control
\cite{xu2024drivegpt4,shao2024lmdrive,sima2024drivelm,tian2024drivevlm}.
Their ability to describe visible road users is promising, but
safety-critical driving requires more than object recognition. A
defensive driver must identify which element constitutes the primary
hazard, estimate whether the risk is immediate or may emerge later,
assess its severity, and select a suitable preventive action.

This distinction is particularly important for \emph{latent hazards}.
A pedestrian partly hidden behind a vehicle, a bus blocking the forward
view, an unstable mixed-traffic interaction, or a roadside crowd may not
represent an immediate collision threat in a single frame. Nevertheless,
the scene may require slowing down, maintaining distance, avoiding an
overtake, or preparing to brake before the danger becomes explicit.
Existing driving-VLM benchmarks increasingly examine corner cases and
model reliability
\cite{chen2025codalm,xie2025drivebench,xing2024autotrust}, but
fine-grained defensive reasoning under weakly structured traffic remains
underexplored.

Bangladesh road scenes present a demanding test environment. They often
contain heterogeneous vehicles, rickshaws and easybikes, CNG
auto-rickshaws, motorcycles, informal pedestrian crossings, roadside
markets, large-vehicle occlusion, damaged surfaces, waterlogging, and
limited lane organization. South Asian datasets such as IDD and IDD-3D
have established the importance of evaluating perception beyond
well-structured Western road environments
\cite{varma2019idd,dokania2023idd3d}. Bangladesh-specific datasets such
as RSUD20K and TFP-BD provide geographically relevant imagery
\cite{zunair2024rsud20k,islam2025tfpbd}, but their original annotations
do not directly evaluate hazard timing, severity, or defensive action
selection.

Geographic mismatch may also encourage unsupported assumptions. A VLM
may infer zebra crossings, formal lane boundaries, sidewalks, traffic
signals, or organized road behavior even when such evidence is absent.
Prior studies demonstrate that multimodal models can hallucinate visual
content and exhibit systematic geographic or contextual biases
\cite{li2023pope,guan2024hallusionbench,huang2025geobias}. Such errors
are especially concerning in driving, where a plausible but visually
ungrounded explanation may conceal unsafe reasoning.

To study these limitations, we introduce \dataset{}, a 500-image
Bangladesh-centric benchmark with structured annotations for hazard presence,
primary hazard category, risk timing, severity, defensive actions, visible evidence,
hallucination traps, and Western-road-bias traps. We evaluate \qwen{}, \internvl{},
and \minicpm{} using a shared zero-shot task definition and structured output schema.

Our results show that binary hazard-presence accuracy is misleading:
a trivial majority-class predictor reaches 89.0\%, essentially matching
all evaluated models. Fine-grained reasoning is substantially harder.
Hazard-category accuracy remains below 40\%, risk-timing accuracy ranges
from 1.6\% to 12.4\%, and exact defensive-action matching remains below
2\%. Moreover, \internvl{}, despite achieving the highest aggregate
severity accuracy, fails to identify any of the 58 high-severity cases.
The models also exhibit category mode collapse and near-zero self-reported recognition
of hallucination- and Western-bias-prone scenes.

Our main contributions are:

\begin{itemize}
    \item We introduce \dataset{}, a 500-image diagnostic benchmark for
    defensive-driving reasoning in Bangladesh mixed-traffic scenes.

    \item We define a structured taxonomy covering twelve locally
    relevant hazard categories, temporal risk, severity, defensive
    actions, visible evidence, and scene-level hallucination and
    Western-road-bias traps.

    \item We provide a controlled zero-shot evaluation of three open
    VLMs using the same 500 images, semantic task definition, output
    schema, and per-image statistical comparisons.

    \item We identify systematic failures in category selection, latent
    hazard anticipation, high-severity recognition, context-specific
    action recommendation, and self-reported trap awareness.
\end{itemize}

\section{Related Work}
\label{sec:related_work}

\subsection{Vision-Language Models for Autonomous Driving}

Recent work has explored language-conditioned driving systems for scene
understanding, explanation, planning, and control. DriveGPT4 combines
multi-frame visual input with language generation and low-level control
prediction, while LMDrive studies language-guided closed-loop driving with
multi-view sensors and natural-language instructions
\cite{xu2024drivegpt4,shao2024lmdrive}. DriveLM organizes perception,
prediction, and planning questions as a graph-structured reasoning task,
and DriveVLM decomposes scene description, analysis, and hierarchical
planning into language-mediated modules
\cite{sima2024drivelm,tian2024drivevlm}. OmniDrive extends this direction
toward 3D grounding and counterfactual reasoning for decision making and
planning \cite{wang2025omnidrive}.

These systems demonstrate the potential of language models for
interpretable driving, but most are designed around full-stack driving,
trajectory prediction, question answering, or closed-loop control. Their
benchmarks are also largely built from datasets such as nuScenes, CARLA,
or BDD-X, whose visual and infrastructural assumptions differ from dense,
weakly structured Bangladesh mixed traffic. In contrast, \dataset{}
isolates image-level defensive reasoning and evaluates whether a model can
identify the primary hazard, distinguish latent from immediate risk,
estimate severity, and recommend an appropriate defensive response.

\subsection{Hazard, Corner-Case, and Reliability Evaluation}

Driving corner cases have increasingly been used to test whether VLMs are
visually grounded and safety-aware. CODA-LM evaluates large VLMs on
self-driving corner cases through general perception, regional
perception, and driving-suggestion tasks \cite{chen2025codalm}.
DriveBench evaluates reliability under clean, corrupted, and text-only
inputs and shows that plausible driving answers may arise from textual
priors or general knowledge rather than visual evidence
\cite{xie2025drivebench}. AutoTrust broadens evaluation to trustfulness,
safety, robustness, privacy, and fairness in driving VLMs
\cite{xing2024autotrust}.

\dataset{} is complementary to these efforts. It does not evaluate
end-to-end control or generic trustworthiness across many attack types.
Instead, it targets a narrower defensive-driving question: whether a VLM
can reason about locally relevant hazards before they become immediate.
Its structured labels separate hazard identity, temporal risk, severity,
and defensive action, enabling diagnosis of failures that may be hidden
by a single overall score or by binary hazard detection. Recent benchmarks examine complementary safety-reasoning capabilities.
DRAMA-X evaluates vulnerable-road-user intent, risk, and ego-action
suggestions, while SCD-Bench studies safety cognition under ambiguous,
misleading, or unsafe driving instructions
\cite{godbole2025dramax,zhang2025scdbench}. NuScenes-SpatialQA and
RoadSceneBench instead emphasize spatial and relational road-scene
reasoning \cite{tian2025nuscenesspatialqa,liu2025roadscenebench}, whereas
INSIGHT improves hazard localization and reasoning through supervised
alignment \cite{chen2025insight}. In contrast, \dataset{} evaluates
zero-shot, image-level defensive reasoning in Bangladesh mixed traffic
with explicit labels for hazard identity, timing, severity, and action.

\subsection{South Asian and Bangladesh Road-Scene Datasets}

Geographic coverage remains a central limitation of autonomous-driving
data. The Indian Driving Dataset (IDD) introduced pixel-level annotations
for unconstrained road environments, while IDD-3D added multi-camera and
LiDAR data for 3D detection and tracking in unstructured Indian traffic
\cite{varma2019idd,dokania2023idd3d}. These datasets established the need
to evaluate perception under heterogeneous vehicle types, weak lane
discipline, dense traffic, and road layouts that differ from those in
widely used Western benchmarks.

For Bangladesh, RSUD20K provides more than 20,000 driving-perspective
images with object-detection annotations for locally relevant road users
and vehicles \cite{zunair2024rsud20k}. TFP-BD contains 23,678 images from
urban Dhaka traffic videos, including irregular pedestrian movement,
different road configurations, and varied environmental conditions
\cite{islam2025tfpbd}. Both datasets are valuable sources of
geographically relevant visual data, but their original annotations
target object detection, traffic flow, or pedestrian analysis rather than
defensive-driving reasoning. \dataset{} builds on their imagery by adding
image-level hazard categories, risk timing, severity, recommended
actions, and scene-level hallucination and geographic-bias traps.

\subsection{Hallucination and Geographic Bias in VLMs}

Large VLMs can generate objects or attributes that are unsupported by the
image. POPE introduced systematic object-hallucination probing, while
HallusionBench evaluates entangled language hallucination, visual
illusion, and logical inconsistency
\cite{li2023pope,guan2024hallusionbench}. Bias benchmarks further show
that multimodal models can exhibit systematic response disparities across
social and geographic contexts \cite{wang2024vlbiasbench}. Recent
geographic evaluation also reports weaker VLM performance for less
wealthy and sparsely represented regions, together with systematic
location over-prediction \cite{huang2025geobias}.

In driving scenes, such failures may appear as unsupported assumptions
about lane markings, zebra crossings, sidewalks, traffic signals, or
other infrastructure that is common in highly represented training
regions but absent from the image. \dataset{} therefore annotates
\emph{hallucination traps} and \emph{Western-road-bias traps}. These labels
describe scene susceptibility, not verified hallucinations in a model's
complete response. Accordingly, our evaluation measures whether a model
recognizes that unsupported reasoning is possible; direct hallucination
measurement would require separate evidence-based human adjudication. Unlike NOPE and DO-Bench, which directly evaluate unsupported object
existence claims and diagnose perceptual versus prior-driven failures
\cite{lovenia2023nope,wang2026dobench}, our trap fields measure whether a
model self-reports that a scene is susceptible to unsupported reasoning.

\section{The BD-HazardVLM Benchmark}
\label{sec:benchmark}

\subsection{Benchmark Scope and Image Sources}

\dataset{} is an image-level diagnostic benchmark for evaluating
defensive driving hazards in Bangladesh road scenes. Table~\ref{tab:benchmark_composition} summarizes the benchmark
composition by source dataset. It contains 500 images, comprising 375 images from
RSUD20K~\cite{zunair2024rsud20k} and 125 images from
TFP-BD~\cite{islam2025tfpbd}. The initial 100-image pilot subset is included within the final
500-image benchmark.

The source datasets provide complementary visual conditions. RSUD20K
contributes road-level scenes containing pedestrians, rickshaws,
easybikes, CNG auto-rickshaws, motorcycles, buses, trucks, private
vehicles, and informal roadside activity. TFP-BD contributes dense
mixed-traffic scenes and broader traffic-flow views. Each benchmark image
is assigned a unique identifier from \texttt{BDHV\_0001} to
\texttt{BDHV\_0500}. All annotations and model predictions are matched
using this identifier rather than source-specific file paths.

The benchmark is designed for image-level reasoning rather than object
detection, segmentation, or trajectory prediction. It therefore provides
scene-level labels describing the most important defensive-driving hazard
and the corresponding safe response.

\begin{table}[!t]
\caption{Composition of the \dataset{} benchmark.}
\label{tab:benchmark_composition}
\centering
\small
\setlength{\tabcolsep}{8pt}
\begin{tabular}{@{}lrr@{}}
\toprule
Source dataset & Number of images & Proportion \\
\midrule
RSUD20K & 375 & 75.0\% \\
TFP-BD  & 125 & 25.0\% \\
\midrule
Total   & 500 & 100.0\% \\
\bottomrule
\end{tabular}
\end{table}

\subsection{Scenario-Driven Image Selection}

Images were selected to maximize scenario diversity rather than to
estimate the natural prevalence of road hazards. The selection process
covered dense mixed traffic, informal pedestrian crossings,
rickshaw/easybike interactions, unpredictable CNG stopping, motorcycle
weaving, large-vehicle occlusion, damaged road surfaces, waterlogging,
roadside markets, animal- or cart-related obstacles, uncontrolled
intersections, low visibility, and scenes without an obvious major
hazard.

The sampling process also considered road type, traffic density, weather,
time of day, visibility, vehicle composition, pedestrian activity, and
the presence of visually competing hazards. This diagnostic selection
strategy intentionally emphasizes difficult cases and should not be
interpreted as a representative estimate of the frequency of hazards
across all Bangladesh roads.

\subsection{Annotation Taxonomy and Schema}

Each image is annotated using a structured image-level schema. The core
evaluation fields are:

\begin{itemize}
    \item \nolinkurl{hazard_present}: whether a major defensive-driving
    hazard is present;

    \item \nolinkurl{primary_hazard_category}: the most important hazard
    affecting the ego driver's decision;

    \item \nolinkurl{risk_timing}: whether the risk is
    \nolinkurl{immediate}, \nolinkurl{latent}, \nolinkurl{both}, or
    \nolinkurl{none};

    \item \nolinkurl{severity}: the estimated consequence level,
    categorized as \nolinkurl{low}, \nolinkurl{medium}, or
    \nolinkurl{high};

   \item \nolinkurl{defensive_actions}: one or more recommended actions
selected from a predefined vocabulary;

\item \nolinkurl{evidence}: a brief description of the visible cues
supporting the annotation;

\item \nolinkurl{hallucination_trap}: whether the scene may encourage
unsupported visual claims;

    \item \nolinkurl{western_bias_trap}: whether the scene may encourage
    unsupported assumptions based on Western road infrastructure.
\end{itemize}

The taxonomy contains twelve mutually exclusive primary hazard
categories: informal pedestrian crossing; rickshaw or easybike hazard;
unpredictable CNG stopping; bus or truck occlusion; motorcycle
wrong-way movement or weaving; road-surface damage; waterlogging;
roadside market or crowd; animal- or cart-related obstacle;
uncontrolled intersection; low visibility; and control scenes without
a major hazard. The machine-readable labels are provided in the
released annotation guidelines.

The defensive-action vocabulary covers slowing down, stopping or yielding,
preparing to brake, maintaining distance, avoiding overtaking, monitoring
occlusion, changing lanes carefully, proceeding cautiously, and taking no
action. Multiple actions are evaluated as an unordered set.

Risk-timing and severity labels are image-level defensive assessments based
on visible scene evidence. They are not physical estimates of time-to-collision or
dynamic collision probability, which would require motion and ego-state
information unavailable from a single frame.

\subsection{Annotation and Quality Control}

Four undergraduate annotators participated in the annotation process,
including three Computer Science and Engineering students and one Physics
student. The 500 images were divided into four disjoint subsets of 125
images, with each annotator labeling one subset using the same annotation
guideline and predefined taxonomy.

Two members of the annotation team---one CSE undergraduate and one Physics
undergraduate---served as reviewers and examined all 500 annotations for
label consistency, taxonomy compliance, evidence quality, and formatting
errors. Ambiguous or inconsistent cases were revisited jointly with the
original annotator and resolved through discussion and consensus. When
consensus remained difficult, the project lead made the final decision
based on visible image evidence and the predefined annotation rules.

This protocol provides complete centralized review of the benchmark.
However, the initial annotations were not produced independently by
multiple annotators for every image. We therefore do not report a formal
inter-annotator agreement statistic and discuss this limitation in
Section~\ref{sec:limitations}. Consequently, scores for subjective attributes such
as risk timing, severity, and defensive-action sets should be interpreted relative
to the adjudicated \dataset{} labels rather than as estimates against a unique
human-consensus ground truth.

\subsection{Dataset Characteristics}

Table~\ref{tab:dataset_statistics} summarizes several important benchmark
characteristics. A total of 445 images contain a major hazard, while 55
are annotated as control scenes. Risk timing is strongly dominated by
latent hazards: 404 images are labeled \texttt{latent}, 41 are labeled
\texttt{immediate}, and 55 are labeled \texttt{none}. No ground-truth
image is labeled \texttt{both}.

\begin{table}[!t]
\caption{Selected statistics of the 500-image benchmark.}
\label{tab:dataset_statistics}
\centering
\small
\setlength{\tabcolsep}{5pt}
\begin{tabular}{@{}lr@{}}
\toprule
Statistic & Count or prevalence \\
\midrule
Hazard-positive images & 445 (89.0\%) \\
Control/no-major-hazard images & 55 (11.0\%) \\
Latent timing & 404 (80.8\%) \\
Immediate timing & 41 (8.2\%) \\
No-risk timing & 55 (11.0\%) \\
Both timing & 0 (0.0\%) \\
Hallucination-trap positive & 324 (64.8\%) \\
Western-bias-trap positive & 285 (57.0\%) \\
\bottomrule
\end{tabular}
\end{table}

The resulting class imbalance is intentional because the benchmark
focuses on difficult hazard reasoning rather than balanced binary
classification. Consequently, raw hazard-presence accuracy is treated as
a secondary metric, while category-level, timing, severity, and
defensive-action performance form the primary evaluation.

\subsection{Release and Licensing}

The original source images will not be redistributed. Instead, the public
release will contain the image manifest, source-image identifiers,
annotations, taxonomy and annotation guidelines, evaluation prompts,
model predictions, parsing utilities, and evaluation scripts. Users will
obtain the original images from their respective dataset repositories and
reconstruct the benchmark using the released manifest.

The annotation, documentation, and model-output layers are intended for
release under the Creative Commons Attribution-NonCommercial 4.0
International license, while the accompanying evaluation and preparation
code is intended for release under the MIT License. The release will
retain source-dataset attribution and licensing information.

\section{Evaluation Protocol}
\label{sec:evaluation}

\subsection{Models and Inference Setup}

We evaluate three open vision-language models:
\qwen{}~\cite{bai2025qwen25vl},
\internvl{}~\cite{zhu2025internvl3}, and
\minicpm{}~\cite{openbmb2026minicpmv46,yao2024minicpmv}. Each model processes the same final set of 500 images in a
zero-shot setting without task-specific training or fine-tuning. All models use greedy decoding with sampling disabled to reduce
stochastic variation.

Model-specific inference settings are summarized in
Table~\ref{tab:model_config}. \qwen{} is loaded using 4-bit NF4
quantization with double quantization and FP16 computation. Its visual
input budget ranges from 200,704 to 401,408 pixels. \internvl{} uses
8-bit quantization and dynamic aspect-ratio-preserving
$448\times448$ image tiling, with at most six tiles and an additional
thumbnail. \minicpm{} is evaluated in FP16 using its native slice-based
visual representation, with 4x downsampling and at most nine slices.

\begin{table}[!t]
\caption{Model-specific inference configurations.}
\label{tab:model_config}
\centering
\small
\setlength{\tabcolsep}{3.5pt}
\begin{tabular}{@{}lccc@{}}
\toprule
Model & Precision & Image processing & Max tokens \\
\midrule
\qwen
& 4-bit NF4
& 200k--401k pixels
& 180 \\

\internvl
& 8-bit
& $448^2$, up to 6 tiles
& 256 \\

\minicpm
& FP16
& 4x, up to 9 slices
& 190 \\
\bottomrule
\end{tabular}
\end{table}

The models retain their native processors and image representations.
Consequently, numerical precision and visual preprocessing are not
identical across models. These differences reflect GPU-memory constraints
and are disclosed when interpreting small performance differences.

\subsection{Prompt and Output Schema}

All models receive the same semantic task instruction, Bangladesh
mixed-traffic context, hazard taxonomy, action vocabulary, and output
schema. The prompt requires the model to rely only on visible image
evidence and explicitly warns against inventing lane markings, traffic
signals, zebra crossings, sidewalks, stop signs, or other unsupported
Western-road infrastructure.

Each response contains eight structured fields:

\begin{quote}
\small
\ttfamily
\raggedright
hazard\_present, primary\_hazard\_category, risk\_timing, severity,\\
defensive\_actions, evidence, possible\_hallucination,\\
possible\_western\_road\_bias
\end{quote}

For control scenes, the required outputs are the control category,
no-risk timing, low severity, and no action. Model outputs are parsed
into the predefined label space. \minicpm{} is allowed one
formatting-only retry when its initial response violates the schema;
the retry provides no additional visual or semantic information.
The evidence field is retained for qualitative auditing and is not
scored as a separate quantitative task.

\subsection{Evaluation Metrics}

Hazard presence is evaluated using accuracy and compared with the
majority-class baseline. Because 89.0\% of the benchmark is
hazard-positive, this binary task is treated as secondary.

For hazard category, risk timing, and severity, we report exact-match
accuracy; for hazard category, we additionally report macro-F1.
Defensive actions are treated as unordered sets and evaluated using exact set match, any-overlap rate, and mean Jaccard similarity.

The model-generated hallucination and Western-bias fields are evaluated
against the corresponding human trap annotations using positive-flag rate
and trap recall. These quantities measure whether a model recognizes that
a scene is susceptible to unsupported reasoning; they do not measure
actual hallucination frequency.

\subsection{Statistical Analysis}

For category, timing, and severity, pairwise model differences are tested
using exact McNemar tests on per-image correctness. Defensive-action
Jaccard scores are compared using paired Wilcoxon signed-rank tests.
Within each task, the three pairwise $p$-values are adjusted using the
Holm procedure. Corrected values below $0.05$ are considered statistically
significant.

\section{Results}
\label{sec:results}

\subsection{Overall Performance}

Table~\ref{tab:main_results} summarizes the core diagnostic performance
of the three models on all 500 benchmark images. No model dominates every
task. \qwen{} achieves the highest hazard-category accuracy,
\internvl{} obtains the highest risk-timing and severity accuracy, and
\minicpm{} achieves the highest defensive-action Jaccard score.

\begin{table}[!t]
\caption{Core performance on the 500-image \dataset{} benchmark.
All values are percentages; higher is better.}
\label{tab:main_results}
\centering
\small
\setlength{\tabcolsep}{3.5pt}
\begin{tabular}{@{}lccccc@{}}
\toprule
Model
& Cat. Acc.
& Cat. F1
& Timing Acc.
& Severity Acc.
& Action Jac. \\
\midrule
\qwen
& \textbf{38.2}
& 9.9
& 9.2
& 58.8
& 22.0 \\

\internvl
& 35.0
& 9.4
& \textbf{12.4}
& \textbf{61.2}
& 25.8 \\

\minicpm
& 27.6
& \textbf{10.5}
& 1.6
& 49.8
& \textbf{29.9} \\
\bottomrule
\end{tabular}
\end{table}

\paragraph{Hazard-presence baseline.}
Raw hazard-presence accuracy ranges from 88.0\% to 89.2\%. However,
445 of the 500 ground-truth images are hazard-positive. A trivial
majority-class classifier that always predicts hazard-present therefore
achieves 89.0\% accuracy, essentially matching all three models.
Consequently, binary hazard presence does not meaningfully discriminate
model capability on this benchmark. The fine-grained category, timing,
severity, and action tasks form the primary evaluation.

\paragraph{Statistical comparisons.}
Using the Holm-adjusted tests defined in
Section~\ref{sec:evaluation}, the category
difference between \qwen{} and \internvl{} is not significant
($p=0.101$), whereas both outperform \minicpm{} ($p<0.001$).
For timing, \internvl{} outperforms \qwen{} ($p=0.005$), and both
outperform \minicpm{} ($p<0.001$). Severity differences are significant
between \internvl{} and \qwen{} ($p=0.036$) and between either model and
\minicpm{} ($p<0.001$). All action-Jaccard pairwise differences are
significant ($p<0.001$).

\subsection{Hazard-Category Recognition}

Hazard-category accuracy remains below 40\% for all evaluated models.
\qwen{} obtains 38.2\%, followed by \internvl{} at 35.0\% and
\minicpm{} at 27.6\%. Their corresponding category macro-F1 scores are
9.9\%, 9.4\%, and 10.5\%, respectively. Thus, \minicpm{} has the lowest
exact accuracy but the highest macro-F1, reflecting broader---although
still weak---coverage across the hazard taxonomy.

\qwen{} exhibits substantial category mode collapse, assigning the
rickshaw/easybike hazard category to 426 of 500 images (85.2\%), although this
category appears in only 205 ground-truth images (41.0\%). Locally salient vehicle
classes therefore often dominate the model's
interpretation of the complete scene,
including cases where the primary hazard is pedestrian behavior,
large-vehicle occlusion, road damage, or an uncontrolled intersection.

\subsection{Risk-Timing and Severity Estimation}

Risk timing is the weakest evaluated capability. \internvl{} obtains
12.4\% accuracy, \qwen{} obtains 9.2\%, and \minicpm{} obtains only
1.6\%. The ground truth contains 404 latent, 41 immediate, and 55
no-risk (\texttt{none}) cases, with no image labeled \texttt{both}.
In contrast, \minicpm{} predicts \texttt{both} for 283 images and
\texttt{immediate} for 217 images, while never predicting
\texttt{latent} or \texttt{none}. It is therefore correct on only
eight images.

A raw-output audit confirms that the 1.6\% \minicpm{} timing score is not
caused by a label-extraction mismatch: every final raw
\texttt{risk\_timing} value exactly matches its parsed value. In 215
initial responses, the model produced the semantically equivalent phrase
``both immediate and latent'' rather than the permitted token
\texttt{both}. The predefined single formatting-repair retry subsequently
returned a schema-valid output. The final distribution therefore reflects
genuine prediction collapse rather than parser loss.

Aggregate severity accuracy is higher, but it conceals safety-critical class
failures. \internvl{} achieves the highest aggregate severity accuracy at 61.2\%,
yet its recall on the 58 high-severity cases is 0.0\%. High-severity recall is only
5.2\% for \qwen{} and 22.4\% for \minicpm{}. Both \internvl{} and \qwen{}
predominantly collapse toward medium severity. \minicpm{} recognizes more
high-severity cases but also overpredicts that class.

\subsection{Defensive-Action Recommendation}

Exact action-set matching is 1.6\% for \qwen{}, 0.4\% for \internvl{},
and 1.0\% for \minicpm{}. Because exact matching requires the complete
predicted action set to equal the reference set, partial-overlap measures
are more informative. The any-overlap rate is 66.6\%, 70.6\%, and 73.2\%,
while mean Jaccard similarity is 22.0\%, 25.8\%, and 29.9\%,
respectively.

The models frequently recommend broadly safe actions such as
\texttt{slow\allowbreak\_down} and
\texttt{prepare\allowbreak\_to\allowbreak\_brake}.
However, they struggle to select context-specific responses such as
\texttt{monitor\allowbreak\_occlusion},
\texttt{avoid\allowbreak\_overtaking},
\texttt{keep\allowbreak\_distance}, or
\texttt{stop\allowbreak\_or\allowbreak\_yield}.
General safety language therefore does not consistently translate into
precise defensive behavior.

\subsection{Hallucination- and Western-Road-Bias Trap Recognition}

Human annotations mark 64.8\% of images as hallucination-prone and 57.0\%
as Western-road-bias-prone, yet all models almost always produce negative
self-reported flags. Table~\ref{tab:trap_flags} quantifies this mismatch.

\begin{table}[!t]
\caption{Ground-truth trap prevalence and unconditional model positive-flag
rates, expressed as percentages. Each model flag rate is computed over all
500 images and may therefore include false positives on trap-negative
images.}
\label{tab:trap_flags}
\centering
\small
\setlength{\tabcolsep}{4pt}
\begin{tabular}{@{}lcccc@{}}
\toprule
Trap
& GT prev.
& Qwen
& InternVL
& MiniCPM \\
\midrule
Hallucination
& 64.8
& 0.0
& 0.4
& 0.6 \\

Western-road bias
& 57.0
& 3.8
& 2.8
& 0.2 \\
\bottomrule
\end{tabular}
\end{table}

For hallucination traps, recall is 0.0\%, 0.3\%, and 0.0\% for \qwen{},
\internvl{}, and \minicpm{}, respectively. For Western-road-bias traps,
recall is 1.1\%, 1.8\%, and 0.0\%, respectively.

These recall values differ from the unconditional flag rates in
Table~\ref{tab:trap_flags} because recall is conditioned only on
ground-truth-positive images, whereas the flag rates include false
positives across all 500 images. For example, \qwen{} produces 19
Western-bias-positive flags, corresponding to a 3.8\% unconditional flag
rate. Only three of these flags occur among the 285 ground-truth-positive
images, yielding 1.1\% recall. Similarly, \minicpm{} produces three
hallucination-trap flags, but all three are false positives, resulting in
a nonzero 0.6\% flag rate and 0.0\% recall.

These results measure self-reported trap recognition rather than
actual hallucination frequency. A positive or negative self-reported
flag does not establish whether the complete textual response contains an
unsupported claim. Measuring actual hallucinations would require separate
human adjudication of the raw model responses using a predefined rubric.

\section{Qualitative Error Analysis}
\label{sec:qualitative}

\begin{figure}[!t]
\centering

\begin{subfigure}[t]{0.48\textwidth}
    \centering
    \includegraphics[width=\linewidth,height=3.0cm,keepaspectratio]
    {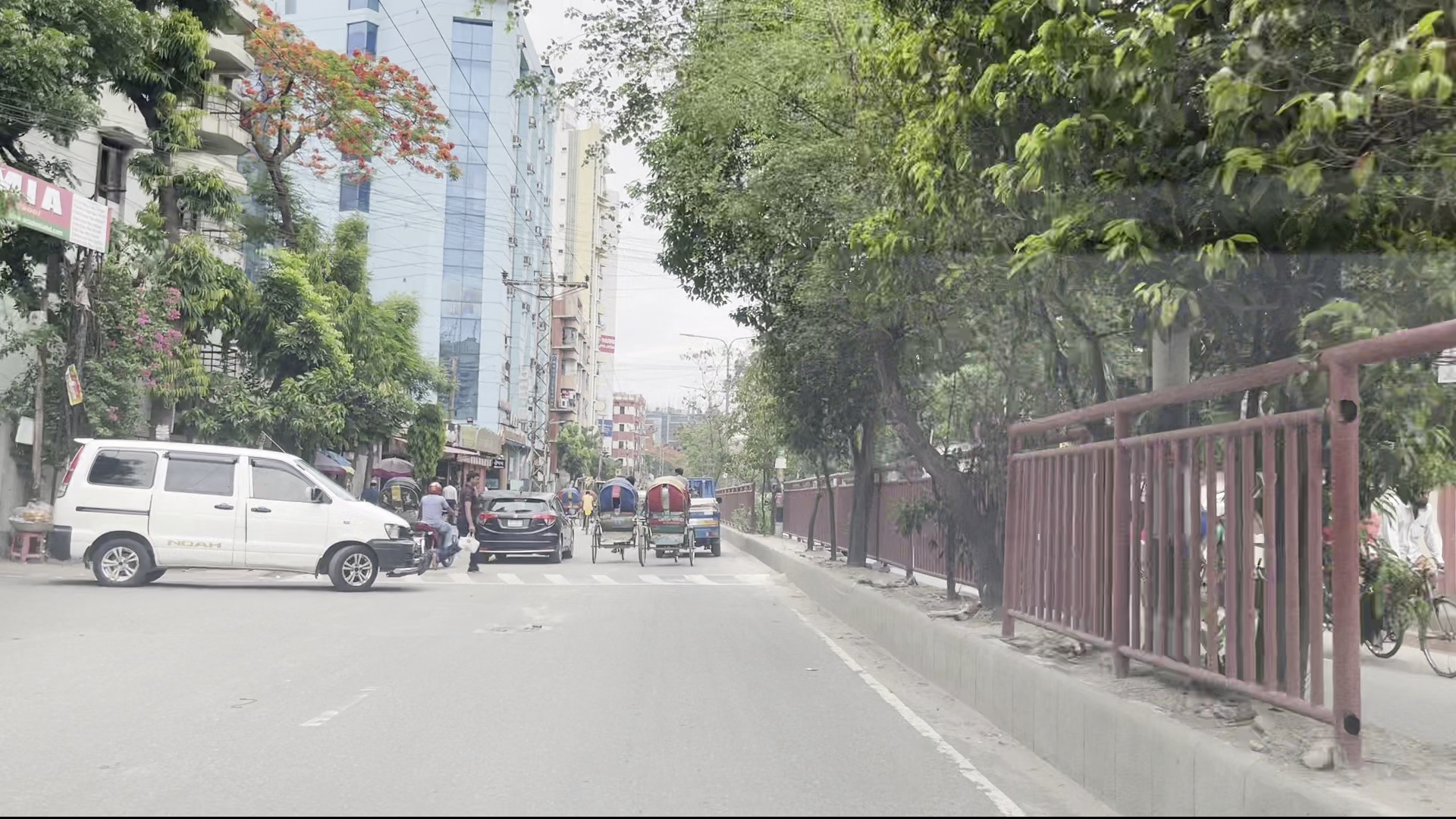}
    \caption{\textbf{GT:} Control, none, low.
    All models predict a hazard.}
\end{subfigure}
\hfill
\begin{subfigure}[t]{0.48\textwidth}
    \centering
    \includegraphics[width=\linewidth,height=3.0cm,keepaspectratio]
    {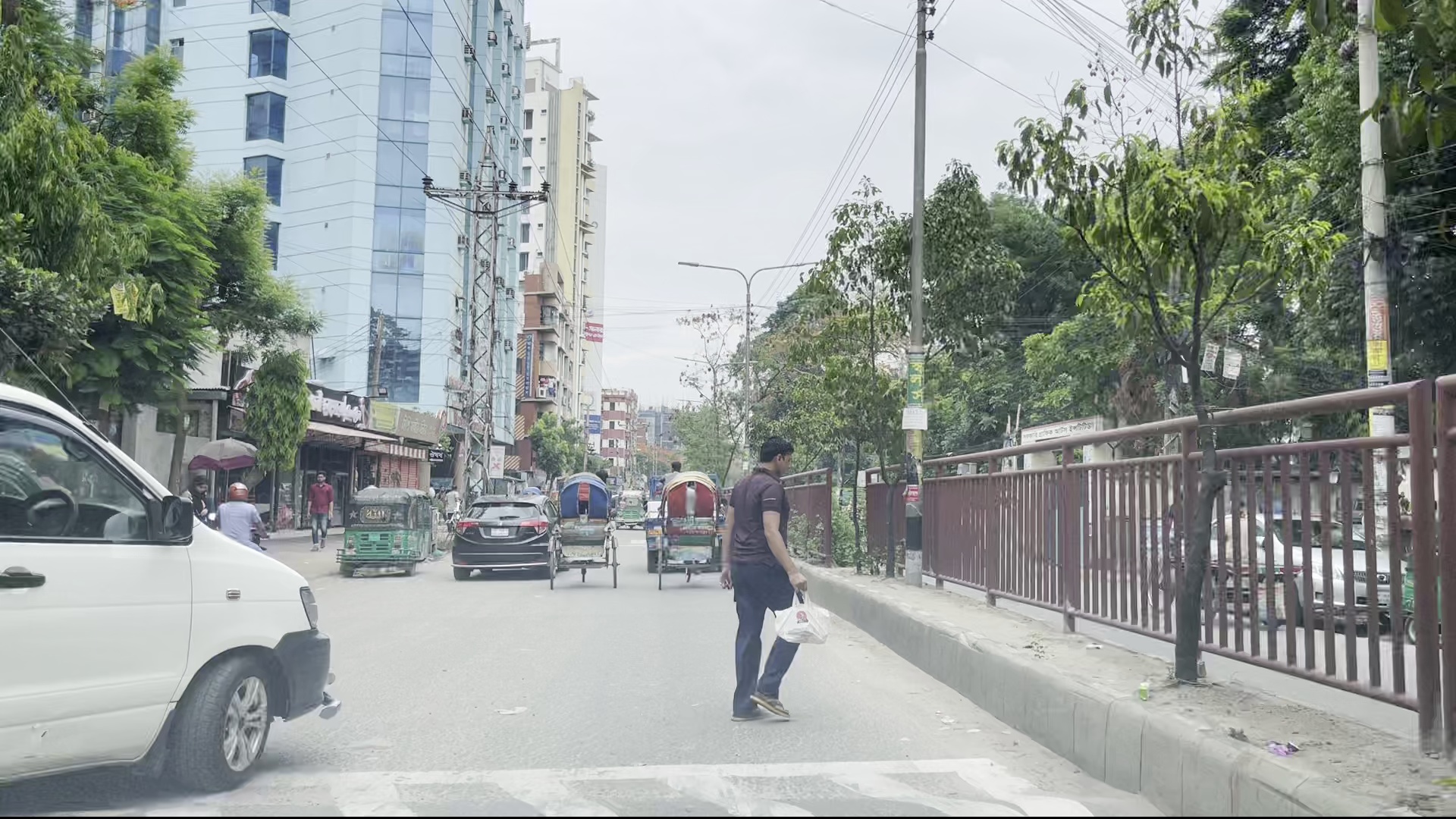}
    \caption{\textbf{GT:} Pedestrian, latent, medium.
    Latent risk becomes immediate or both.}
\end{subfigure}

\vspace{2mm}

\begin{subfigure}[t]{0.48\textwidth}
    \centering
    \includegraphics[width=\linewidth,height=3.0cm,keepaspectratio]
    {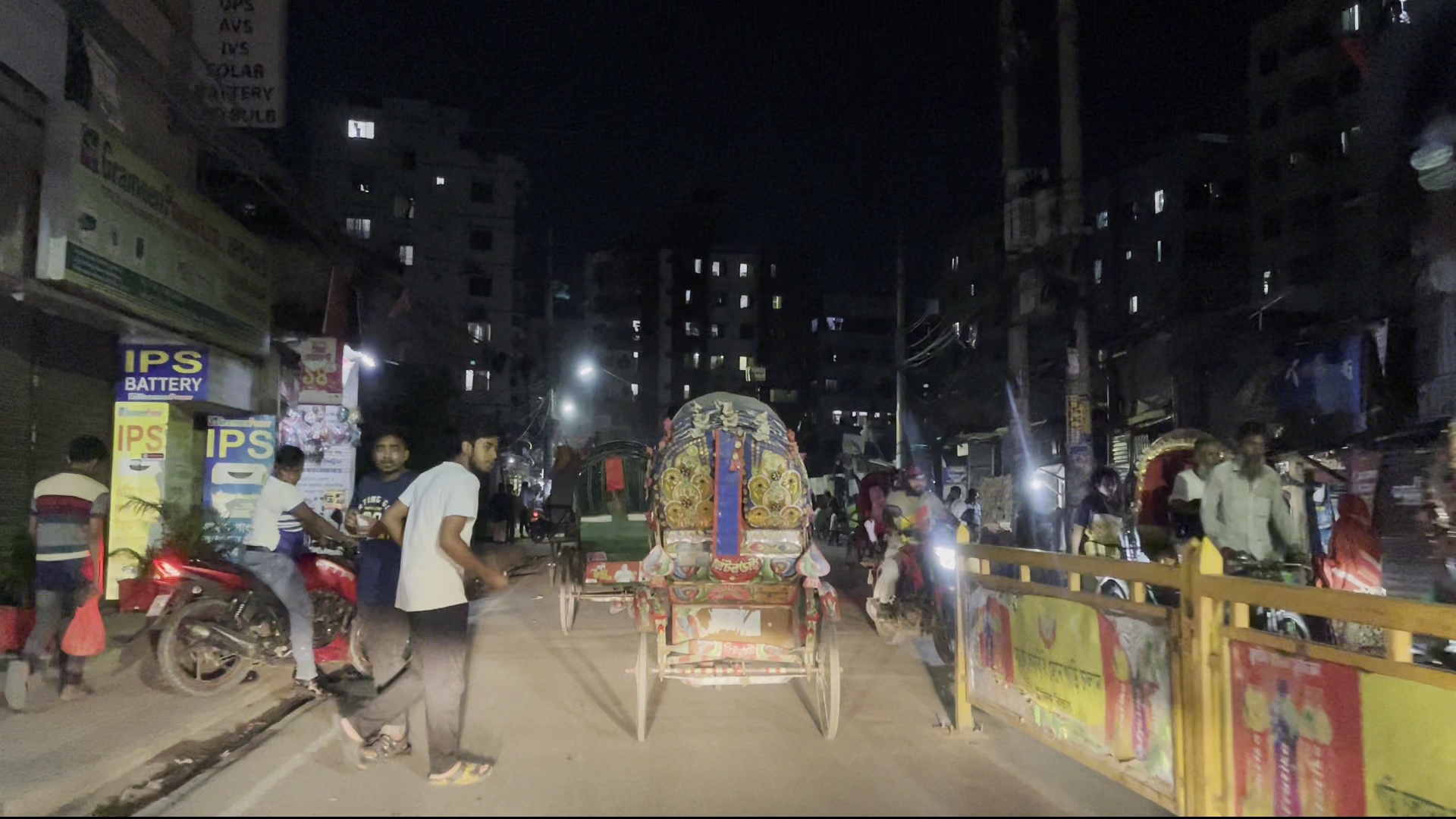}
    \caption{\textbf{GT:} Pedestrian, immediate, high.
    Models focus on a rickshaw and predict medium.}
\end{subfigure}
\hfill
\begin{subfigure}[t]{0.48\textwidth}
    \centering
    \includegraphics[width=\linewidth,height=3.0cm,keepaspectratio]
    {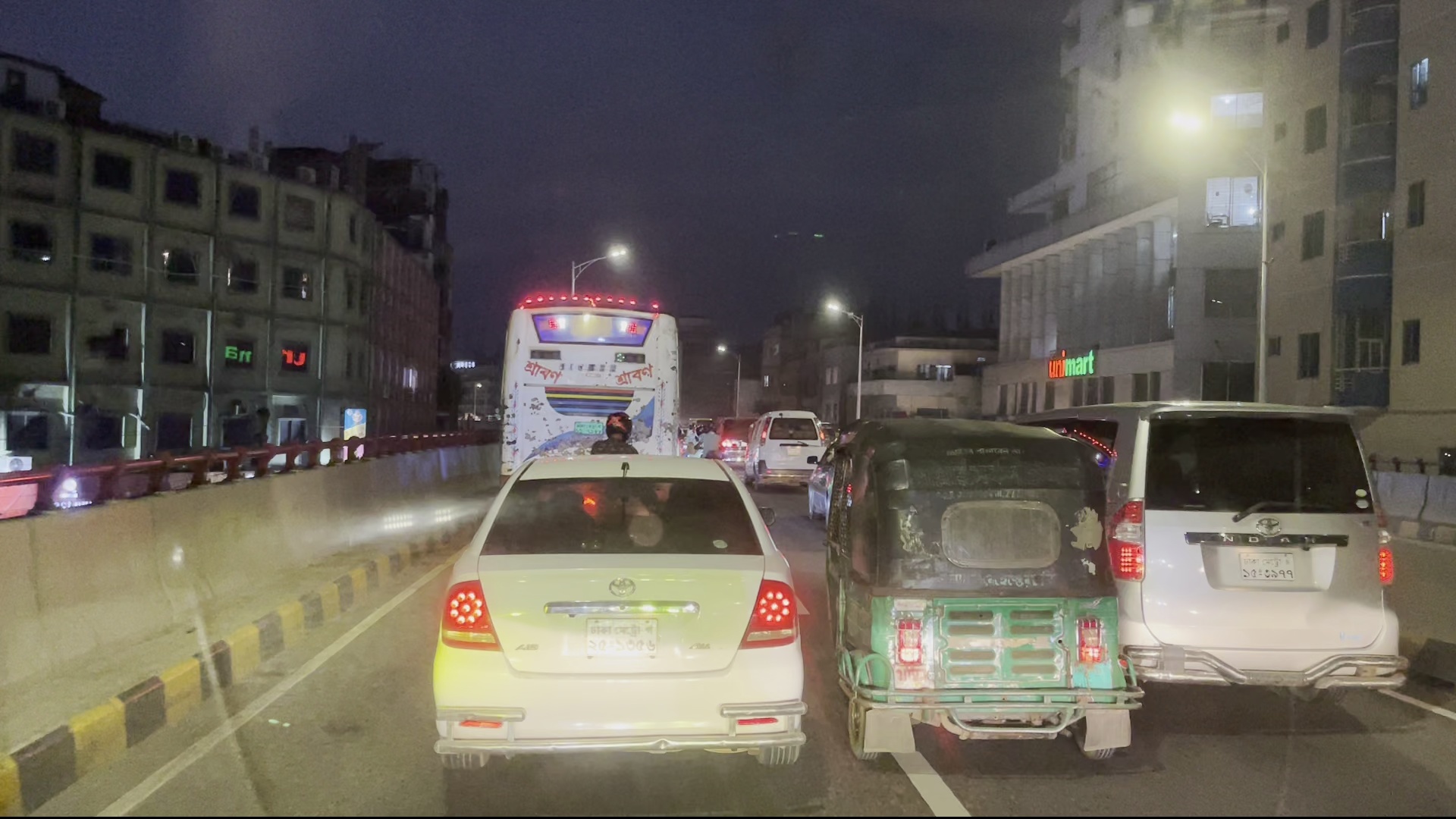}
    \caption{\textbf{GT:} Bus/truck occlusion, latent, medium.
    Correct category still yields incorrect timing.}
\end{subfigure}

\caption{Representative failure cases from \dataset{}. The examples
illustrate false hazard activation, weak latent-risk anticipation,
high-severity underestimation, and inconsistent reasoning across output
fields.}
\label{fig:qualitative_cases}
\end{figure}

The four cases in Fig.~\ref{fig:qualitative_cases} illustrate recurrent failure
patterns. In the control scene in Fig.~\ref{fig:qualitative_cases}(a),
all models activate a hazard response despite the absence of a major
defensive-driving threat. In Fig.~\ref{fig:qualitative_cases}(b), the
pedestrian hazard is partially recognized, but its latent timing is
converted into an immediate or combined risk.

Figure~\ref{fig:qualitative_cases}(c) shows high-severity
underestimation: the models focus on a visually salient rickshaw rather
than prioritizing the vulnerable pedestrian. In
Fig.~\ref{fig:qualitative_cases}(d), \minicpm{} identifies the correct
bus/truck-occlusion category but still predicts incorrect timing and
severity. These cases show that correct object or category recognition
does not guarantee coherent defensive-driving reasoning across category,
timing, severity, and action fields.

\section{Limitations and Ethical Considerations}
\label{sec:limitations}

\paragraph{Scope and annotation.}
\dataset{} is a focused 500-image diagnostic benchmark rather than a
representative sample of all Bangladesh roads. Images were selected for
scenario diversity, producing substantial class imbalance, particularly
for hazard presence and risk timing. The benchmark also uses individual
frames and therefore cannot directly evaluate motion, trajectories, or
time-to-collision. Risk-timing and severity labels are thus image-level assessments
from visible evidence; unobserved ego speed, acceleration, and object motion can
change the appropriate dynamic interpretation of the same frame. Each image
received one initial annotation and centralized review, but not independent
duplicate annotation. Because both reviewers were also annotators, each reviewed
their own 125-image subset. We therefore do not report formal inter-annotator
agreement, and the benchmark does not yet establish an independent
human-agreement ceiling for the more subjective attributes.

\paragraph{Taxonomy and model comparability.}
The predefined taxonomy covers recurring Bangladesh mixed-traffic hazards
but simplifies scenes containing multiple interacting risks by assigning
one primary category. Moreover, although all models process the same 500
images and receive the same semantic task, they retain different native
processors and numerical-precision settings because of GPU-memory
constraints. Small performance differences should therefore be interpreted
with care. Accordingly, the reported results characterize these three evaluated
open models under the disclosed inference configurations and should not be
interpreted as establishing equivalent failure rates for larger or proprietary
VLMs.

\paragraph{Hallucination interpretation.}
The hallucination- and Western-bias-related outputs are model-generated
self-reports. Their comparison with human trap labels measures whether a
model recognizes that unsupported reasoning is possible; it does not
measure the actual hallucination rate of the complete response. Direct
measurement would require separate evidence-based human adjudication.

\paragraph{Release and responsible use.}
The original images will not be redistributed. We will release source
identifiers, the image manifest, annotations, taxonomy documentation,
prompts, model outputs, and evaluation utilities, while requiring users to
obtain images from the original repositories and follow their licenses.
The annotations are intended for research on multimodal reasoning and
driving safety, not identification, surveillance, or safety-critical
deployment. \dataset{} is not a validated vehicle-safety component.

\section{Conclusion}
\label{sec:conclusion}

We introduced \dataset{}, a 500-image Bangladesh-centric benchmark for
image-level defensive-driving reasoning across hazard category, risk
timing, severity, defensive actions, and reasoning traps.

Across three VLMs, category accuracy remains below 40\%, timing accuracy
ranges from 1.6\% to 12.4\%, and exact action-set matching remains below
2\%. \internvl{} misses all 58 high-severity cases despite its leading
aggregate severity accuracy, while \qwen{} exhibits severe category
collapse. All models show weak latent-risk anticipation and near-zero
self-reported trap recognition. For these three evaluated open VLMs, the results
show that coarse hazard recognition does not ensure reliable fine-grained
defensive reasoning. Future work will add duplicate annotation, video evaluation,
broader baselines, and subgroup analysis.

\bibliographystyle{splncs04}
\bibliography{main}

@article{xu2024drivegpt4,
  author  = {Xu, Zhenhua and Zhang, Yujia and Xie, Enze and Zhao, Zhen and Guo, Yong and Wong, Kwan-Yee K. and Li, Zhenguo and Zhao, Hengshuang},
  title   = {DriveGPT4: Interpretable End-to-End Autonomous Driving via Large Language Model},
  journal = {IEEE Robotics and Automation Letters},
  volume  = {9},
  number  = {10},
  pages   = {8186--8193},
  year    = {2024},
  doi     = {10.1109/LRA.2024.3440097}
}

@inproceedings{shao2024lmdrive,
  author    = {Shao, Hao and Hu, Yuxuan and Wang, Letian and Song, Guanglu and Waslander, Steven L. and Liu, Yu and Li, Hongsheng},
  title     = {LMDrive: Closed-Loop End-to-End Driving with Large Language Models},
  booktitle = {Proceedings of the IEEE/CVF Conference on Computer Vision and Pattern Recognition},
  pages     = {15120--15130},
  year      = {2024}
}

@inproceedings{sima2024drivelm,
  author    = {Sima, Chonghao and Renz, Katrin and Chitta, Kashyap and Chen, Li and Zhang, Hanxue and Xie, Chengen and Beisswenger, Jens and Luo, Ping and Geiger, Andreas and Li, Hongyang},
  title     = {DriveLM: Driving with Graph Visual Question Answering},
  booktitle = {Computer Vision -- ECCV 2024},
  pages     = {256--274},
  year      = {2024},
  doi       = {10.1007/978-3-031-72943-0_15}
}

@article{tian2024drivevlm,
  author  = {Tian, Xiaoyu and Gu, Junru and Li, Bailin and Liu, Yicheng and Wang, Yang and Zhao, Zhiyong and Zhan, Kun and Jia, Peng and Lang, Xianpeng and Zhao, Hang},
  title   = {DriveVLM: The Convergence of Autonomous Driving and Large Vision-Language Models},
  journal = {arXiv preprint arXiv:2402.12289},
  year    = {2024}
}

@inproceedings{wang2025omnidrive,
  author    = {Wang, Shihao and Yu, Zhiding and Jiang, Xiaohui and Lan, Shiyi and Shi, Min and Chang, Nadine and Kautz, Jan and Li, Ying and Alvarez, Jose M.},
  title     = {OmniDrive: A Holistic Vision-Language Dataset for Autonomous Driving with Counterfactual Reasoning},
  booktitle = {Proceedings of the IEEE/CVF Conference on Computer Vision and Pattern Recognition},
  pages     = {22442--22452},
  year      = {2025}
}

@inproceedings{chen2025codalm,
  author    = {Chen, Kai and Li, Yanze and Zhang, Wenhua and Liu, Yanxin and Li, Pengxiang and Gao, Ruiyuan and Hong, Lanqing and Tian, Meng and Zhao, Xinhai and Li, Zhenguo and Yeung, Dit-Yan and Lu, Huchuan and Jia, Xu},
  title     = {Automated Evaluation of Large Vision-Language Models on Self-Driving Corner Cases},
  booktitle = {Proceedings of the IEEE/CVF Winter Conference on Applications of Computer Vision},
  pages     = {7806--7815},
  year      = {2025},
  doi       = {10.1109/WACV61041.2025.00759}
}

@inproceedings{xie2025drivebench,
  author    = {Xie, Shaoyuan and Kong, Lingdong and Dong, Yuhao and Sima, Chonghao and Zhang, Wenwei and Chen, Qi Alfred and Liu, Ziwei and Pan, Liang},
  title     = {Are VLMs Ready for Autonomous Driving? An Empirical Study from the Reliability, Data and Metric Perspectives},
  booktitle = {Proceedings of the IEEE/CVF International Conference on Computer Vision},
  pages     = {6585--6597},
  year      = {2025}
}

@article{xing2024autotrust,
  author  = {Xing, Shuo and Hua, Hongyuan and Gao, Xiangbo and Zhu, Shenzhe and Li, Renjie and Tian, Kexin and Li, Xiaopeng and Huang, Heng and Yang, Tianbao and Wang, Zhangyang and Zhou, Yang and Yao, Huaxiu and Tu, Zhengzhong},
  title   = {AutoTrust: Benchmarking Trustworthiness in Large Vision Language Models for Autonomous Driving},
  journal = {arXiv preprint arXiv:2412.15206},
  year    = {2024}
}

@inproceedings{varma2019idd,
  author    = {Varma, Girish and Subramanian, Anbumani and Namboodiri, Anoop and Chandraker, Manmohan and Jawahar, C. V.},
  title     = {IDD: A Dataset for Exploring Problems of Autonomous Navigation in Unconstrained Environments},
  booktitle = {IEEE Winter Conference on Applications of Computer Vision},
  pages     = {1743--1751},
  year      = {2019},
  doi       = {10.1109/WACV.2019.00190}
}

@inproceedings{dokania2023idd3d,
  author    = {Dokania, Shubham and Hafez, A. H. Abdul and Subramanian, Anbumani and Chandraker, Manmohan and Jawahar, C. V.},
  title     = {IDD-3D: Indian Driving Dataset for 3D Unstructured Road Scenes},
  booktitle = {Proceedings of the IEEE/CVF Winter Conference on Applications of Computer Vision},
  pages     = {4482--4491},
  year      = {2023}
}

@inproceedings{zunair2024rsud20k,
  author    = {Zunair, Hasib and Khan, Shakib and Hamza, A. Ben},
  title     = {RSUD20K: A Dataset for Road Scene Understanding in Autonomous Driving},
  booktitle = {IEEE International Conference on Image Processing},
  pages     = {708--714},
  year      = {2024},
  doi       = {10.1109/ICIP51287.2024.10648203}
}

@article{islam2025tfpbd,
  author  = {Islam, Mohammad Manzurul and Das, Aritra and Shams, Karib and Hasan, Md. Rakibul and Hasan, Kazi Ferdous and Rashid, Mohammad Rifat Ahmmad and Chowdhury, Abdullahi and Ali, Md. Sawkat and Islam, Maheen and Shahjalal, Mohammed and Masum, Saleh},
  title   = {TFP-BD: An Image Dataset for Traffic Flow and Pedestrian Movement Analysis on Bangladeshi Urban Roads},
  journal = {Data in Brief},
  volume  = {59},
  pages   = {111398},
  year    = {2025},
  doi     = {10.1016/j.dib.2025.111398}
}

@inproceedings{li2023pope,
  author    = {Li, Yifan and Du, Yifan and Zhou, Kun and Wang, Jinpeng and Zhao, Wayne Xin and Wen, Ji-Rong},
  title     = {Evaluating Object Hallucination in Large Vision-Language Models},
  booktitle = {Proceedings of the 2023 Conference on Empirical Methods in Natural Language Processing},
  pages     = {292--305},
  year      = {2023},
  doi       = {10.18653/v1/2023.emnlp-main.20}
}

@inproceedings{guan2024hallusionbench,
  author    = {Guan, Tianrui and Liu, Fuxiao and Wu, Xiyang and Xian, Ruiqi and Li, Zongxia and Liu, Xiaoyu and Wang, Xijun and Chen, Lichang and Huang, Furong and Yacoob, Yaser and Manocha, Dinesh and Zhou, Tianyi},
  title     = {HallusionBench: An Advanced Diagnostic Suite for Entangled Language Hallucination and Visual Illusion in Large Vision-Language Models},
  booktitle = {Proceedings of the IEEE/CVF Conference on Computer Vision and Pattern Recognition},
  pages     = {14375--14385},
  year      = {2024}
}

@article{wang2024vlbiasbench,
  author  = {Wang, Sibo and Cao, Xiangkui and Zhang, Jie and Yuan, Zheng and Shan, Shiguang and Chen, Xilin and Gao, Wen},
  title   = {VLBiasBench: A Comprehensive Benchmark for Evaluating Bias in Large Vision-Language Models},
  journal = {arXiv preprint arXiv:2406.14194},
  year    = {2024}
}

@inproceedings{huang2025geobias,
  author    = {Huang, Jingyuan and Huang, Jen-tse and Liu, Ziyi and Liu, Xiaoyuan and Wang, Wenxuan and Zhao, Jieyu},
  title     = {AI Sees Your Location---But With a Bias Toward the Wealthy World},
  booktitle = {Proceedings of the 2025 Conference on Empirical Methods in Natural Language Processing},
  pages     = {18019--18039},
  year      = {2025},
  doi       = {10.18653/v1/2025.emnlp-main.910}
}

@article{bai2025qwen25vl,
  author  = {Bai, Shuai and Chen, Keqin and Liu, Xuejing and Wang, Jialin and Ge, Wenbin and Song, Sibo and Dang, Kai and Wang, Peng and Wang, Shijie and Tang, Jun and Zhong, Humen and Zhu, Yuanzhi and Yang, Mingkun and Li, Zhaohai and Wan, Jianqiang and Wang, Pengfei and Ding, Wei and Fu, Zheren and Xu, Yiheng and Ye, Jiabo and Zhang, Xi and Xie, Tianbao and Cheng, Zesen and Zhang, Hang and Yang, Zhibo and Xu, Haiyang and Lin, Junyang},
  title   = {Qwen2.5-VL Technical Report},
  journal = {arXiv preprint arXiv:2502.13923},
  year    = {2025}
}

@article{zhu2025internvl3,
  author  = {Zhu, Jinguo and Wang, Weiyun and Chen, Zhe and Liu, Zhaoyang and Ye, Shenglong and Gu, Lixin and Tian, Hao and Duan, Yuchen and Su, Weijie and Shao, Jie and others},
  title   = {InternVL3: Exploring Advanced Training and Test-Time Recipes for Open-Source Multimodal Models},
  journal = {arXiv preprint arXiv:2504.10479},
  year    = {2025}
}

@misc{openbmb2026minicpmv46,
  author       = {{OpenBMB}},
  title        = {MiniCPM-V 4.6 Model Card},
  year         = {2026},
  howpublished = {\url{https://huggingface.co/openbmb/MiniCPM-V-4.6}},
  note         = {Accessed July 2026}
}

@article{yao2024minicpmv,
  author  = {Yao, Yuan and Yu, Tianyu and Zhang, Ao and Wang, Chongyi and Cui, Junbo and Zhu, Hongji and Cai, Tianchi and Li, Haoyu and Zhao, Weilin and He, Zhihui and others},
  title   = {MiniCPM-V: A GPT-4V Level MLLM on Your Phone},
  journal = {arXiv preprint arXiv:2408.01800},
  year    = {2024}
}

@article{godbole2025dramax,
  author  = {Godbole, Mihir and Gao, Xiangbo and Tu, Zhengzhong},
  title   = {DRAMA-X: A Fine-grained Intent Prediction and Risk Reasoning Benchmark for Driving},
  journal = {arXiv preprint arXiv:2506.17590},
  year    = {2025}
}

@article{zhang2025scdbench,
 author = {Zhang, Enming and Gong, Peizhe and Dai, Xingyuan and
          Huang, Min and Lv, Yisheng and Miao, Qinghai},
  title   = {Evaluation of Safety Cognition Capability in Vision-Language Models for Autonomous Driving},
  journal = {arXiv preprint arXiv:2503.06497},
  year    = {2025}
}

@article{tian2025nuscenesspatialqa,
  author  = {Tian, Kexin and Mao, Jingrui and Zhang, Yunlong and Jiang, Jiwan and Zhou, Yang and Tu, Zhengzhong},
  title   = {NuScenes-SpatialQA: A Spatial Understanding and Reasoning Benchmark for Vision-Language Models in Autonomous Driving},
  journal = {arXiv preprint arXiv:2504.03164},
  year    = {2025}
}

@article{liu2025roadscenebench,
  author  = {Liu, Xiyan and Wang, Han and Wang, Yuhu and Cai, Junjie and
             Cao, Zhe and Yang, Jianzhong and Lu, Zhen},
  title   = {{RoadSceneBench}: A Lightweight Benchmark for Mid-Level Road Scene Understanding},
  journal = {arXiv preprint arXiv:2511.22466},
  year    = {2025}
}

@article{chen2025insight,
  author = {Chen, Dianwei and Zhang, Zifan and Cheng, Lei and
          Liu, Yuchen and Yang, Xianfeng Terry},
  title   = {INSIGHT: Enhancing Autonomous Driving Safety through Vision-Language Models on Context-Aware Hazard Detection and Edge Case Evaluation},
  journal = {arXiv preprint arXiv:2502.00262},
  year    = {2025}
}

@article{lovenia2023nope,
  author  = {Lovenia, Holy and Dai, Wenliang and Cahyawijaya, Samuel and Ji, Ziwei and Fung, Pascale},
  title   = {Negative Object Presence Evaluation (NOPE) to Measure Object Hallucination in Vision-Language Models},
  journal = {arXiv preprint arXiv:2310.05338},
  year    = {2023}
}

@article{wang2026dobench,
  author  = {Wang, JiYang and Chen, Jiawei and Xiao, Mengqi and Cheng, Yu and Li, Yangfu and Yin, Zhaoxia},
  title   = {DO-Bench: An Attributable Benchmark for Diagnosing Object Hallucination in Vision-Language Models},
  journal = {arXiv preprint arXiv:2604.22822},
  year    = {2026}
}

\end{document}